\documentclass[10pt,twocolumn,letterpaper]{article}

\usepackage{cvpr}              

\usepackage{graphicx}
\usepackage{amsmath}
\usepackage{amssymb}
\usepackage{booktabs}
\usepackage{makecell}
\usepackage{bbm}
\usepackage{overpic}
\usepackage{color}
\usepackage{epstopdf}
\usepackage{amsthm}
\usepackage[ruled]{algorithm2e}
\usepackage{amsfonts}
\usepackage{multirow,array,bm}
\usepackage{helvet} 
\usepackage{courier}  
\usepackage{caption}
\usepackage[accsupp]{axessibility}
\usepackage[pagebackref,breaklinks,colorlinks]{hyperref}

\usepackage[capitalize]{cleveref}
\crefname{section}{Sec.}{Secs.}
\Crefname{section}{Section}{Sections}
\Crefname{table}{Table}{Tables}
\crefname{table}{Tab.}{Tabs.}

\def\cvprPaperID{3186} 
\def\confName{CVPR}
\def\confYear{2022}

\begin{document}

\title{Dense Learning based Semi-Supervised Object Detection}

\author{Binghui Chen$^{1}$, Pengyu Li$^{1}$, Xiang Chen$^{1}$, Biao Wang$^{1}$, Lei Zhang$^{2}$, Xian-Sheng Hua$^{1}$\\
$^{1}$ Alibaba Group, $^{2}$ The Hong Kong Polytechnic University\\
{\tt\small chenbinghui@bupt.cn, lipengyu007@gmail.com, xchen.cx@alibaba-inc.com, wangbiao225@foxmail.com}\\
{\tt\small cslzhang@comp.polyu.edu.hk, huaxiansheng@gmail.com}
}
\maketitle

\begin{abstract}
Semi-supervised object detection (SSOD) aims to facilitate the training and deployment of object detectors with the help of a large amount of unlabeled data. Though
various self-training based and consistency-regularization based SSOD methods have been proposed, most of them are anchor-based detectors, ignoring the fact that in many real-world applications anchor-free detectors are more demanded. In this paper, we intend to bridge this gap and propose a DenSe Learning (DSL) based anchor-free SSOD algorithm. Specifically, we achieve this goal by introducing several novel techniques, including an Adaptive Filtering strategy for assigning multi-level and accurate dense pixel-wise pseudo-labels, an Aggregated Teacher for producing stable and precise pseudo-labels, and an uncertainty-consistency-regularization term among scales and shuffled patches for improving the generalization capability of the detector. Extensive experiments are conducted on MS-COCO and PASCAL-VOC, and the results show that our proposed DSL method records new state-of-the-art SSOD performance, surpassing existing methods by a large margin. Codes can be found at \textcolor{blue}{https://github.com/chenbinghui1/DSL}.

\end{abstract}

\section{Introduction}
\label{sec:intro}
The recent rapid development of object detection (OD) methods \cite{paa-eccv2020,carion2020end,sun2021sparse} largely owes to the availability of large-scale and well-annotated datasets, such as MS-COCO benchmark \cite{lin2014microsoft}. With the increasing demand for more powerful and accurate detection models, the need to collect and label more data also increases. However, manually labeling the class labels and bounding-boxes for large-scale datasets is a very expensive and tedious job, which is not cost-effective in practical applications. As a remedy, semi-supervised \cite{sohn2020simple,yang2021interactive} and self-supervised \cite{liu2020self} OD algorithms, which aim to employ the large amount of unlabeled data to improve the performance of OD, have been attracting much attention in recent years. In this paper, we focus on the semi-supervised objection detection (SSOD) methods.

\begin{figure}[!t]
  \centering
  \includegraphics[width=1\linewidth]{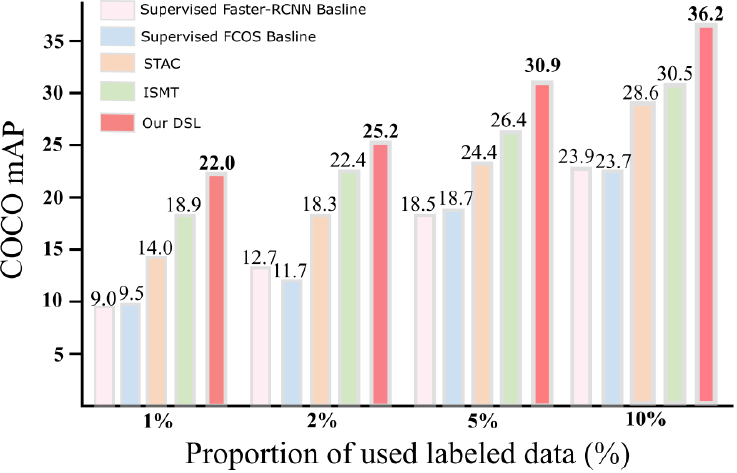}\\
  \vspace{-0em}
  \caption{The SSOD performance comparisons between the proposed anchor-free based DSL and anchor-based methods STAC \cite{sohn2020simple} and ISMT \cite{yang2021interactive}. One can observe that anchor-based detector Faster-RCNN \cite{ren2015faster} and anchor-free based detector FCOS \cite{tian2019fcos} have similar baseline performance under the supervised settings, while our proposed DSL achieves the state-of-the-art SSOD performance, outperforming the existing methods by a large margin.}\label{fig_intro}
  \vspace{-1em}
\end{figure}


The current state-of-the-art SSOD methods are pseudo-label based approaches \cite{sohn2020simple,liu2021unbiased,yang2021interactive,zhou2021instant}, while most of them are based on a two-stage anchor-based detector such as Faster-RCNN \cite{ren2015faster}. Specifically, they first use a teacher model to generate  pseudo-labels for unlabeled images and then train a two-stage anchor-based detector with both labeled and unlabeled images. However, in real-world applications, the one-stage anchor-free based detectors (\eg, FCOS \cite{tian2019fcos}) are more attractive and practical since they are much easier and efficient to be deployed on resource limited devices without heavy pre/post-processing except NMS. Different from Faster-RCNN, the learning of FCOS is established on dense feature predictions; that is, each pixel is directly supervised by the corresponding label. Without the help of predefined anchors and multiple refinements of the predictions, the learning of anchor-free based detectors requires more careful guidance, especially under the SSOD settings. Unfortunately, few works on anchor-free SSOD have been reported, and how to handle the dense pseudo-labels predicted by anchor-free detectors remains a challenging problem.

To address the above mentioned challenges, in this paper we propose a DenSe Learning (DSL) algorithm for anchor-free SSOD \footnote{In this paper, we employ FCOS \cite{tian2019fcos} as our baseline detector.}. Specifically, to perform careful label guidance for dense learning, we first present an Adaptive Filtering (AF) strategy to partition pseudo-labels into three fine-grained parts, including background, foreground, and ignorable regions. Then we refine these pseudo-labels by using a MetaNet so as to remove the classification false-positives, which have higher prediction scores but are actually false predictions in category. Considering that the correctness of pseudo-labels determines the performance of SSOD models, we introduce an Aggregated Teacher (AT) to further enhance the stability and quality of the estimated pseudo-labels. Moreover, to improve the model generalization capability, we learn from shuffled image patches and regularize the uncertainty of dense feature maps to make them consistent among image scales. The main contributions of this paper are summarized as follows:

\begin{itemize}
  \item A simple yet effective DenSe Learning (DSL) method is developed to improve the utilization of large-scale unlabelled data for SSOD. To our best knowledge, this is the first anchor-free method for SSOD.
  \item An Adaptive Filtering (AF) strategy is proposed to assign fine-grained pseudo-labels to each pixel; an Aggregated Teacher (AT) is introduced to enhance the stability and quality of estimated pseudo-labels; and learning from shuffled patches and uncertainty-consistency-regularization among scales are employed to improve the model generalization performance.
\end{itemize}

Extensive experiments conducted on MS-COCO \cite{lin2014microsoft} and PASCAL-VOC \cite{everingham2010pascal} demonstrate that the proposed DSL method achieves significant performance improvements over existing state-of-the-art SSOD methods.

\section{Related Work}
\textbf{Semi-Supervised Learning for Image Classification}.
Recently, semi-supervised learning (SSL) has achieved significant progress in image classification with the rapid development of deep learning techniques. SSL aims to employ a large amount of unlabeled data to learn robust and discriminative classification boundaries. Specifically, self-ensembling is used in \cite{laine2016temporal} to stabilize the learning targets for unlabeled data. A new measure of local smoothness of the conditional label distribution is proposed in \cite{miyato2018virtual} for improving the SSL learning performance.  Mean teacher is employed in \cite{tarvainen2017mean} to produce accurate labels instead of label ensembles. Generally speaking, the above consistency-based methods apply perturbations to the input image and then minimize the differences between their output predictions. These methods have proved to be effective at smoothing the feature manifold, and consequently improving the generalization performance of models. There are also some other techniques targeting at utilizing the unlabeled data to improve image classification, including self-training \cite{chen2016weakly,li2019learning,xie2020self,lee2013pseudo}, data augmentation \cite{berthelot2019mixmatch,sohn2020fixmatch} and so on.

Though many SSL methods have been proposed for image classification, it is not a trivial work to transfer them to the task of object detection due to the complex architectural design and multi-task learning (classification and regression) nature of object detectors.

\textbf{Object Detection} is a fundamental task in computer vision. Current CNN-based object detectors can be categorized into anchor-based and anchor-free methods. Faster R-CNN \cite{ren2015faster} is a well-known and representative two-stage anchor-based detector. It consists of a region proposal network (RPN) and a region-wise prediction network (R-CNN) for detecting objects. Many works \cite{cai2016unified,cai2018cascade,lee2019me,li2019scale,bell2016inside,tian2018learning} have been proposed to improve the performance of Faster RCNN. For anchor-free object detection, the state-of-the-art methods \cite{redmon2016you,huang2015densebox,tian2019fcos,liu2019high,kong2020foveabox} mostly regard the center (\eg, the center point or part) of an object as a foreground to define positives, and then predict the distances from positives to the four sides of the object bounding box (BBox). For example, FCOS \cite{tian2019fcos} takes all the pixels inside the BBox as positives, and uses these four distances and a centerness score to detect objects. CSP \cite{liu2019high} defines only the center point of the object box as positive to detect pedestrians with fixed aspect ratio. FoveaBox \cite{kong2020foveabox} regards pixels in the middle part of object as positives and learns four distances to perform detection. Without the need to set anchors, anchor-free detectors are much easier and more flexible to be deployed in real applications.

\begin{figure*}[!t]
  \centering
  \includegraphics[width=1\linewidth]{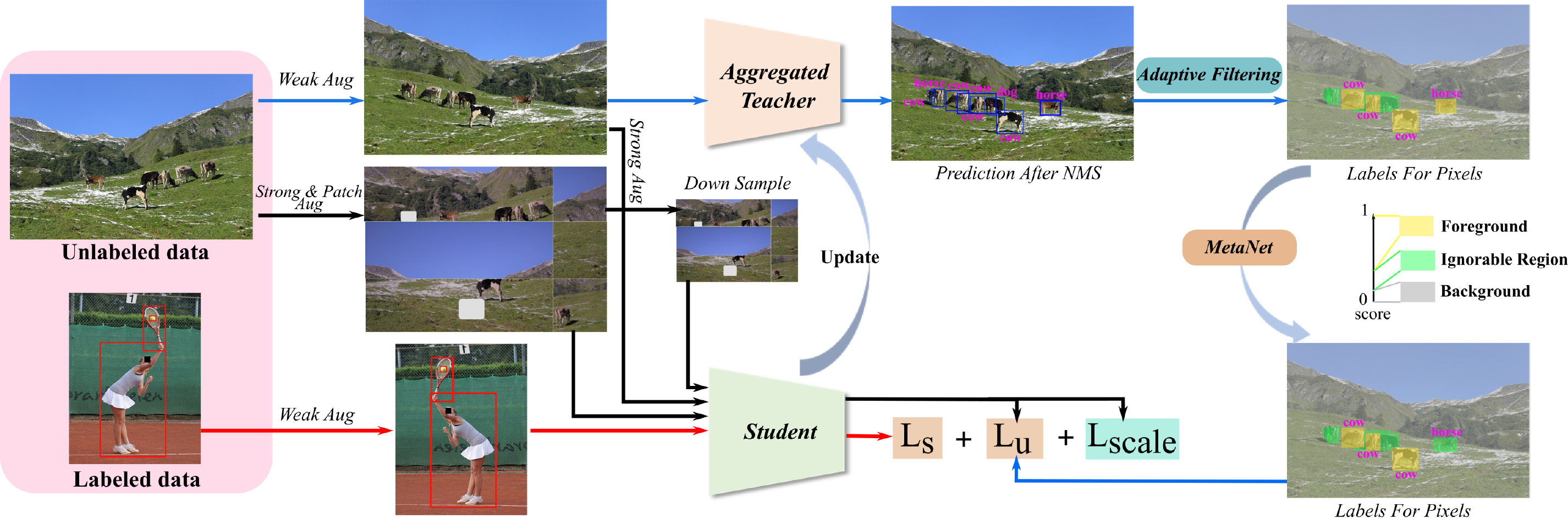}\\
  \vspace{-0.1em}
  \caption{The pipeline of our proposed DenSe Learning (DSL) based SSOD method. The training data contain both labeled and unlabeled images. During each training iteration, a teacher model is employed to produce pseudo-labels for weakly augmented unlabeled images. In anchor-free based detectors like FCOS\cite{tian2019fcos}, each spatial location of the dense predictions will be assigned with one label, and the model performance is sensitive to noisy pseudo-labels. To alleviate this problem, an Adaptive Filtering strategy is proposed to split the pseudo-labels into three types, including background, foreground and ignorable regions. Moreover, there exist some false positive cases, which have higher scores but are obviously wrong predictions. Thus, a MetaNet is proposed to refine these cases. To improve the model generalization capability, unlabeled images are patch-shuffled and consistency regularizations are applied on these images with different scales. For improving the stability and quality of pseudo-labels, the teacher model is updated by the student models via aggregation, called Aggregated Teacher. After obtaining the fine-grained pixel-wise pseudo-labels, the detector can be optimized by the final loss, which is the sum of $L_{s},L_{u}$ and $L_{scale}$.}\label{fig_pipeline}
  \vspace{-0.5em}
\end{figure*}

\textbf{Semi-Supervised Object Detection (SSOD)}. SSOD aims to improve the performance of object detectors by using larger-scale unlabeled data. Since the manual annotation of object labels is very expensive, producing pseudo-labels for unlabeled data is very attractive. In \cite{zoph2020rethinking,radosavovic2018data,song2022prnet++}, the pseudo-labels are produced by ensembling the predictions from different data augmentations. STAC \cite{sohn2020simple} uses both weak and strong augmentations for model training, where strong augmentations are only applied to unlabeled data while weak augmentations are used to produce stable pseudo-labels. UBA \cite{liu2021unbiased} employs the EMA teacher \cite{tarvainen2017mean} for producing more accurate pseudo-labels. ISMT \cite{yang2021interactive} fuses the current pseudo-labels with history labels via NMS, and uses multiple detection heads to improve the accuracy of pseudo-labels. Instant-Teaching \cite{zhou2021instant} combines more powerful augmentations like Mixup and Mosaic into the training stage. Humble-Teacher \cite{tang2021humble} uses plenty of proposals and soft pseudo-labels for the unlabeled data. Certainty-aware pseudo-labels are tailored in \cite{li2021rethinking} for object detection. E2E \cite{xu2021end} uses a soft teacher mechanism for training with the unlabeled data. Almost all the above methods are built upon anchor-based detectors, \eg, Faster RCNN, which are not convenient to deploy in real applications with limited resources. Therefore, in this work we develop, for the first time to our best knowledge, an anchor-free SSOD method.
\section{Methods}
\subsection{Preliminary}
For the convenience of expression, we first provide some notations for the SSOD task. Suppose that we have two sets of data, a labeled set $\mathcal{X}=\{X_{i}|^{N_{l}}_{i=1}\}$ and an unlabeled set $\mathcal{U}=\{U_{i}|^{N_{u}}_{i=1}\}$, where $N_{l}$ and $N_{u}$ are the number of labeled and unlabeled images, respectively, and $N_{u}\gg N_{l}$. Each labeled image has annotations of category $p^{*}\in[0,C-1]$ ($C$ is the number of foreground classes) and annotations of bounding box (BBox) $t^{*}$. In an image, each region annotated by BBox and class label is called an instance. Without loss of generality, we take the anchor-free FCOS \cite{tian2019fcos} detector as our baseline, which is composed of a ResNet50 \cite{he2016deep} backbone, an FPN \cite{lin2017feature} neck and a dense head. To use both labeled and unlabeled data for training, the overall loss can be defined as follows:
\begin{equation}\label{eq_overloss}
  L=L_{s}+\alpha L_{u}
\end{equation}
where $L_{s}$ and $L_{u}$ denote supervised loss and unsupervised loss, respectively, and $\alpha$ is the hyper-parameter to control the contribution of unlabeled data.

Both of the supervised and unsupervised losses are normalized by the corresponding number of positive pixels in each mini-batch as follows:
\begin{small}
\begin{align}\label{eq_FCOS}
  L_{s}=\frac{1}{N_{pos}}\sum_{i}\sum_{h,w}(&L_{cls}(X_{i,h,w})+\mathbbm{1}_{\{p^{*}_{h,w}\in[0,C-1]\}}L_{reg}(X_{i,h,w})\nonumber\\
  +&\mathbbm{1}_{\{p^{*}_{h,w}\in[0,C-1]\}}L_{center}(X_{i,h,w}))
\end{align}
\vspace{-1em}
\begin{align}
  L_{u}=\frac{1}{N_{pos}}\sum_{i}\sum_{h,w}(&L_{cls}(U_{i,h,w})+\mathbbm{1}_{\{\bar{p}^{*}_{h,w}\in[0,C-1]\}}L_{reg}(U_{i,h,w})\nonumber\\
  +&\mathbbm{1}_{\{\bar{p}^{*}_{h,w}\in[0,C-1]\}}L_{center}(U_{i,h,w}))
\end{align}
\end{small}

\noindent
where $N_{pos}$ means the number of positive pixels in one mini-batch, $X_{i,h,w}$ means the predicted vector at spatial location $(h,w)$ from the $i^{th}$ image, $\bar{p}^{*}_{h,w}$ is the corresponding estimated pseudo-labels at location $(h,w)$. $L_{cls},L_{reg}$ and $L_{center}$ are the default losses used in FCOS \cite{tian2019fcos}. $\mathbbm{1}_{\{\cdot\}}$ is the indicator function, which outputs $1$ if condition $\{\cdot\}$ is satisfied and $0$ otherwise.

In this paper, we propose a \textbf{\emph{DenSe Learning}} (DSL) algorithm for bridging the gap between SSOD and anchor-free detector. The pipeline of our DSL method is illustrated in Figure \ref{fig_pipeline}. It is mainly composed of an Adaptive Filtering (AF) strategy, a MetaNet, an Aggregated Teacher (AT) and an Uncertainty-Consistency regularization term, which are introduced in detail in the following sections.

\subsection{Adaptive Filtering Strategy}
\begin{figure}[!t]
  \centering
  \includegraphics[width=0.7\linewidth]{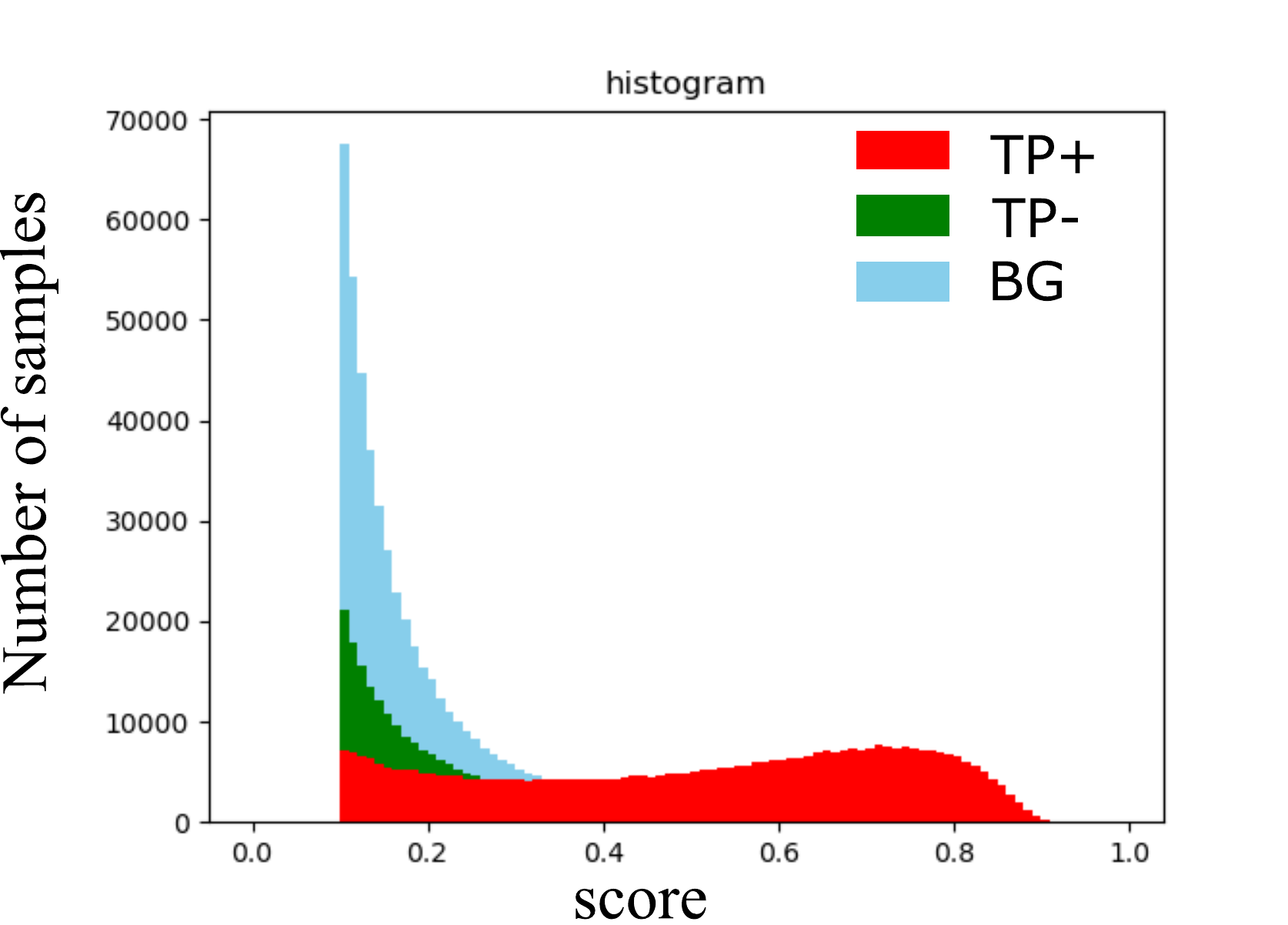}\\
  \vspace{-0.5em}
  \caption{The distributions of TP+, TP- and BG when using $10\%$ labeled data on COCO. `TP+' means that the estimated instance has the same class ID as the ground-truth (GT) and the IOU of BBox is above $0.5$. `TP-' means that the estimated instance has the same class ID as GT but the IOU of BBox is below $0.5$. `BG' means that the estimated instance belongs to the background or has wrong class ID.}\label{fig_dd}
  \vspace{-1em}
\end{figure}

The FCOS \cite{tian2019fcos} detector reduces the dependency on predefined anchors by introducing dense pixel-wise supervision. Though this is helpful for the easy deployment in actual applications, the performance of the model is sensitive to the quality of pixel-wise labels. Because the predicted pseudo-labels in SSOD will have noise no matter how powerful the detector is, the pixel-wise supervision for FCOS should be treated prudently. To this end, we propose an Adaptive Filtering (AF) strategy to elaborately handle the pseudo-labels for dense learning.

To exploit the unlabeled data, we need to assign a pseudo-label for each pixel in the output dense tensor. As shown in Figure \ref{fig_dd}, however, we can see that the TP+, TP- and BG instances coexist with each other, and their distributions are much more complex. If we simply use a single threshold to define foreground and background, many instances will be assigned with wrong labels, resulting in heavy noise and damaging the learning of an accurate detector. For example, if we set a relatively higher threshold $0.4$ to define the positive instances, there will be many TP+ and TP- wrongly assigned to the background. Conversely, if we set a relatively lower threshold $0.1$ to define the background instances, there will be many BG instances wrongly assigned to the foreground. Therefore, we propose to use multiple thresholds $\{\tau_{1},\tau_{2}\}$ to partition the estimated instances into three parts: background, ignorable region and foreground:
\vspace{-0.3em}
\begin{equation}\label{eq_pl}
\bar{p}_{h,w}^{*}=\left\{
\begin{aligned}
&Foreground:[0,\cdots,C-1]& & p_{h,w}>=\tau_{2}, \\
&Ignorable~Region:[-1] & & \tau_{1}<p_{h,w}<\tau_{2},\\
&Background:[C] & & p_{h,w}<=\tau_{1}.
\end{aligned}
\right.
\end{equation}
where $p_{h,w}$ is the predicted score at location $(h,w)$ (If not specified, it is the product of classification score and centerness score), and $\bar{p}_{h,w}^{*}$ is the corresponding pseudo-label. Different from foreground and background regions, we ignore the gradients computation and propagation for ignorable regions as:
\begin{small}
\begin{align}
\label{eq_ignore}
  L_{u}=\frac{1}{N_{pos}}\sum_{i}\sum_{h,w}(&\mathbbm{1}_{\{\bar{p}^{*}_{h,w}\geq0\}}L_{cls}(U_{i,h,w})+\mathbbm{1}_{\{\bar{p}^{*}_{h,w}\in[0,C-1]\}}\nonumber\\
L_{reg}(U_{i,h,w})+&\mathbbm{1}_{\{\bar{p}^{*}_{h,w}\in[0,C-1]\}}L_{center}(U_{i,h,w})).
\end{align}
\end{small}

\noindent
$\tau_{1}$ in Eq. \ref{eq_pl} is used to filter out the background and thus it is relatively easy to set. We set $\tau_{1}=0.1$ throughout our experiments. $\tau_{2}$ is employed to filter out the foreground and it is harder to set for different classes. We propose to use a class-adaptive $\tau_{2}^{k}$ instead of a fixed $\tau_{2}$:
\begin{equation}\label{eq_ada}
  \tau_{2}^{k} = (\frac{\sum_{h,w}\mathbbm{1}_{\{\bar{p}_{h,w}^{*}==k\}}p_{h,w}}{N_{pos}})^{\beta}\tau,
\end{equation}
where $\tau_{2}^{k}$ is the threshold for the $k^{th}$ class, $\beta=0.7$ is used
to control the degree of focus on tail-classes, and $\tau=0.35$ is used as a fixed reference threshold. 

\textbf{Remarks}: Different from those anchor-based detectors, anchor-free detectors will predict each pixel as either background or foreground, and compute gradients for all of them. However, for unlabeled data, instances with scores within interval $[\tau_{1},\tau_{2}^{k}]$ are noisy and confusing, and treating them as either foreground or background will degrade the detection performance. Therefore, in anchor-free SSOD we should explicitly set multiple fine-grained thresholds to identify not only the background and foreground but also the ignorable regions. The proposed AF strategy can well handle this problem and assign fine-grained and multi-level labels to the dense pixels, as illustrated in Figure. \ref{fig_pipeline}. We experimentally demonstrate that the AF strategy is very important for anchor-free SSOD.

\subsection{MetaNet}
\begin{figure}[!t]
  \centering
  \includegraphics[width=1\linewidth]{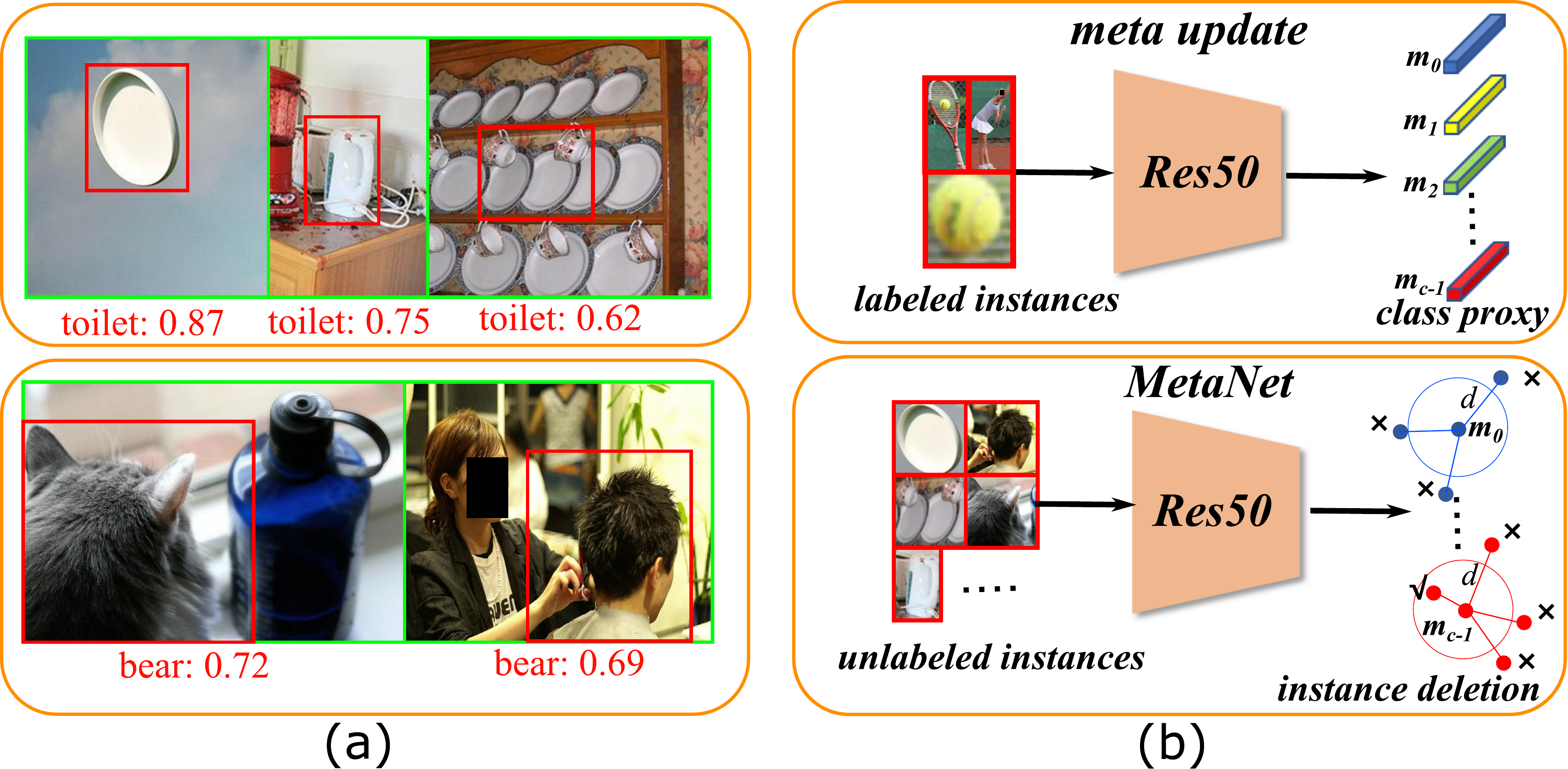}\\
  \vspace{-0.2em}
  \caption{(a) The estimated classification-false-positive instances which have high scores but are obvious false predictions in category. (b) Our proposed MetaNet for refining the pseudo-labels of instances. `$\surd$' and `$\times$' mean reservation and deletion, \emph{resp}.}\label{fig_meta}
  \vspace{-1em}
\end{figure}

Though AF has the ability to improve the quality of pseudo-labels for dense learning, there still exist some classification-false-positive instances, which have high scores but are obvious false predictions, as shown in Figure \ref{fig_meta}(a). In order to handle these instances, we resort to using a MetaNet, as shown in Figure \ref{fig_meta}(b). We use a ResNet50 to implement the MetaNet. Before DSL training, we first pass all the labeled instances into the MetaNet and compute the following class-wise proxies $m_{k}$:
\begin{equation}\label{eq_meta}
  m_{k}=\frac{\sum_{i}f_{i,k}}{N_{k}},
\end{equation}
\noindent
where $f_{i,k}$ is the $1$-D feature vector of the $i^{th}$ instance belonging to the $k^{th}$ class, $N_{k}$ is the number of instances of the $k^{th}$ class. After obtaining the class-wise proxies, we refine the pseudo-labels by computing the cosine distance between the feature vector of the unlabeled instance and the corresponding class proxy vector. If the distance is smaller than a threshold $d=0.6$, we will change the label `Foreground' of this instance to the label `Ignorable Region'.

\textbf{Remarks:} MetaNet is employed to rectify the predicted foreground class labels of those error-prone instances. It only performs the meta update step and thus can work in a plug-and-play manner. The computation of MetaNet only involves the class proxy update on the labeled instances without gradient back-propagation, and thus it is fast and the cost is negligible compared with the training of DSL. With the help of stable class proxies, we can successfully remove many classification-false-positive instances.

\subsection{Aggregated Teacher}
\begin{figure}[!t]
  \centering
  \includegraphics[width=1\linewidth]{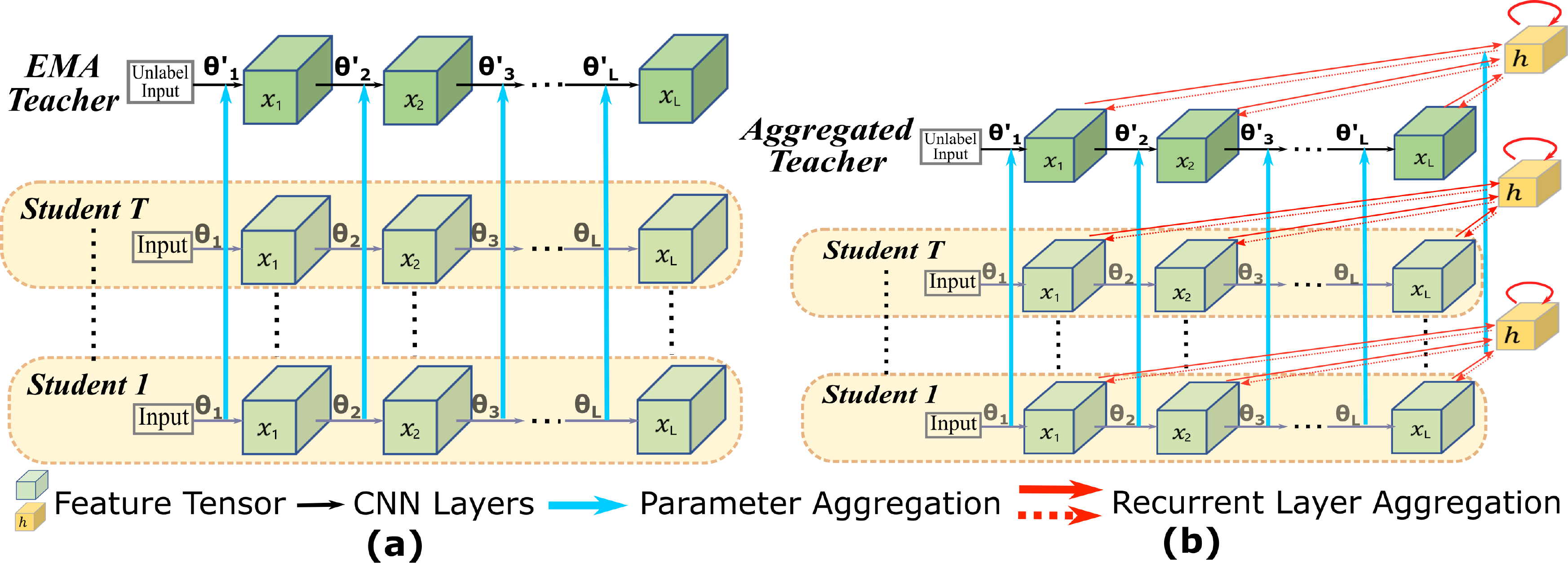}\\
  \vspace{-0.5em}
  \caption{The illustration of (a) EMA Teacher and (b) our Aggregated Teacher. EMA teacher performs aggregation only over parameters, while our Aggregated teacher performs aggregation over both parameters and layers.}\label{fig_at}
  \vspace{-.5em}
\end{figure}

In pseudo-label based methods, the stability and quality of the predicted pseudo-labels are important to the final performance. Therefore, almost all the existing anchor-based methods \cite{liu2021unbiased,yang2021interactive,xu2021end,tang2021humble,li2021rethinking} employ an EMA Teacher to improve the quality of pseudo-labels for the unlabeled data. As illustrated in Figure \ref{fig_at}(a), EMA is usually performed in following manner:
\begin{equation}\label{eq_ema}
  \theta^{'t}=\epsilon\theta^{'t-1}+(1-\epsilon)\theta^{t},
\end{equation}
\noindent
where $\epsilon$ is a smoothing hyperparameter, $t$ means the iteration, $\theta$ and $\theta^{'}$ are parameters of the student and teacher models, respectively.

EMA update aims to obtain a more stable and powerful teacher model via the ensemble of students. However, such an update in Eq. \ref{eq_ema} might still be coarse and weak because it only aggregates parameters in the same layer at different iterations, without considering the correlation across layers. To further enhance the capability of teacher model, motivated by the dense aggregation mechanism \cite{huang2017densely,yu2018deep,zhao2021recurrence}, we introduce an Aggregated Teacher (AT), which performs not only parameter aggregation across time but also recurrent layer aggregation across layers, as illustrated in Figure \ref{fig_at}(b). Specifically, for parameter aggregation, we still adopt the existing EMA update as in Eq. \ref{eq_ema}. While for layer aggregation, to avoid the problem of heavy parameter, we follow the recurrent learning \cite{zhao2021recurrence,hochreiter1997long,lin1996learning} and use a recurrent layer aggregation mechanism as bellow:
\begin{align}
  x_{l+1}&=\theta_{l+1}[x_{l}+h_{l}]+x_{l}\label{eq_res_rla},\\
  h_{l+1}&=g_{2}[g_{1}[\theta_{l+1}[x_{l}+h_{l}]]+h_{l}]\label{eq_rla},
\end{align}
\noindent
where $x_{l}$ is the $l^{th}$ layer's tensor in CNN and $\theta_{l}$ denotes the corresponding convolution parameters. $h_{l}$ is the hidden state tensor for the $l^{th}$ layer, and $h_{1}$ is initialized with zero. $g_{1}$ and $g_{2}$ are the corresponding $1\times1$ and $3\times3$ Conv layers used for recurrent computing, which are parameter-shared across the adjacent layers within the same stage. $*[\cdot]$ indicates the convolution operation between input tensor `$\cdot$' and parameter `$*$'. By using the recurrent mechanism, the number of introduced parameters is negligible. One can see from Eq. \ref{eq_res_rla} that it will degrade to the default residual unit of ResNet when the hidden state $h_{l-1}$ is removed. In other words, the recurrent layer aggregation can be easily applied to the current residual CNN models. Moreover, since neck and heads in the detector are very shallow, we only perform layer aggregation over the backbone.

\textbf{Remarks}: Since the parameter aggregation in EMA Teacher treats each layer independently, the relationship between layers might be destroyed during aggregation, and thus one aggregated layer may not work well with the adjacent ones. Therefore, layer aggregation is considered in our model. By explicitly using the hidden state to connect the current layer with the previous layers, the knowledge propagation will be more stable and accurate. Moreover, the shared recurrent layers impose regularization over the propagated information. Compared with EMA Teacher, the Aggregated Teacher is able to produce more stable and accurate pseudo-labels for dense learning.

\subsection{Uncertainty Consistency}
\begin{figure}[!t]
  \centering
  \includegraphics[width=0.9\linewidth]{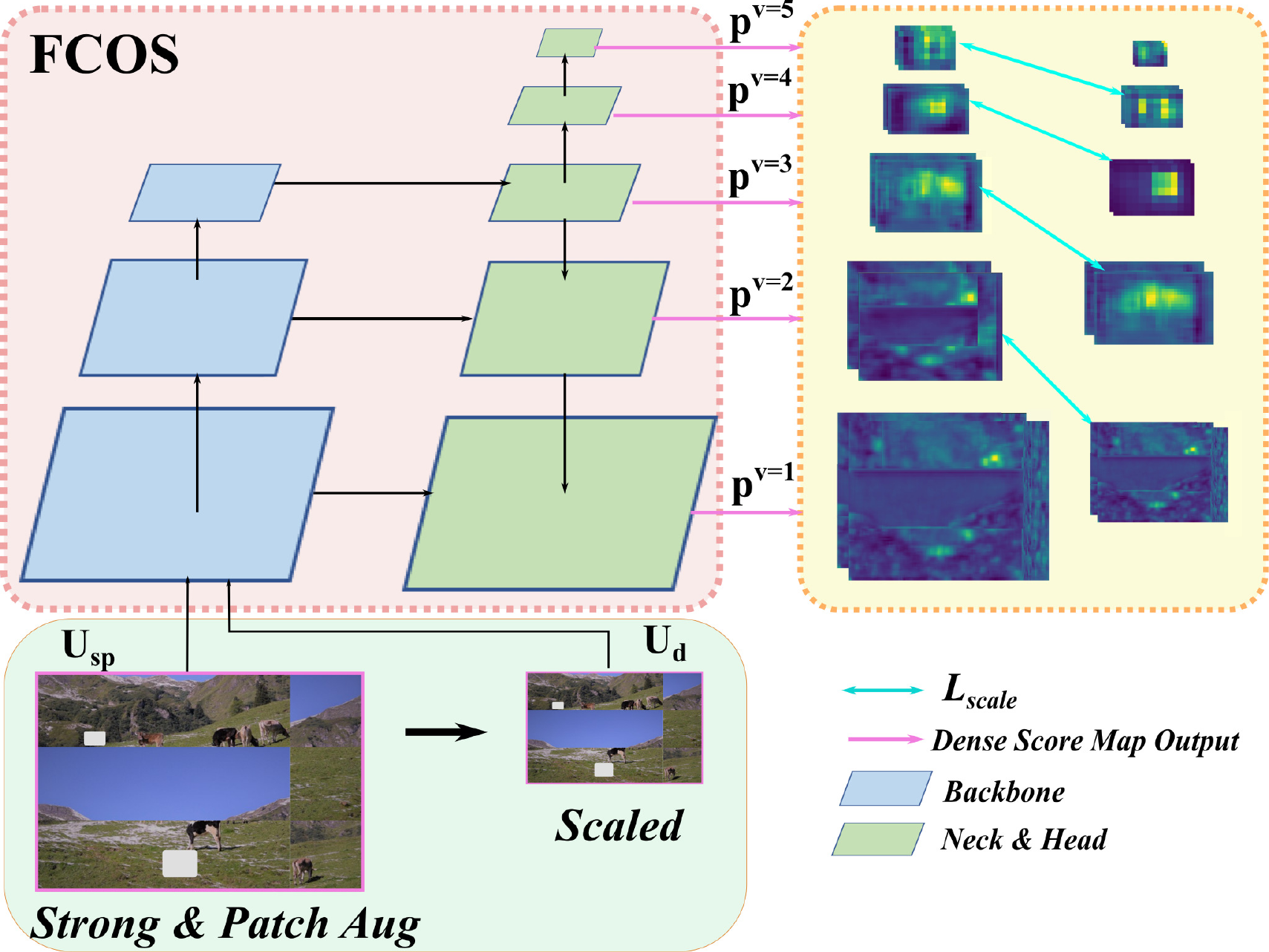}\\
  \caption{Illustration of the uncertainty consistency regularization among scales. The input images come from the same unlabeled image $U_{i}$.}\label{fig_si}
  \vspace{-.5em}
\end{figure}

By using the proposed AF, MetaNet and AT, the dense pixel-wise pseudo-labels can be obtained to supervise the learning of SSOD models by optimizing the loss $L_{u}$. In order to further improve the generalization capability of the SSOD model, we propose to regularize the uncertainty consistency over the unlabeled images. From Figure \ref{fig_si}, one can see that the input consists of a pair of images: Strong \& Patch Augmented image ($U_{sp}$) and the corresponding Down-sampled image ($U_{d}$). The downsampling ratio is set to $r=2$ in producing $U_{d}$. By patch shuffle augmentation, we randomly crop an image into several parts along the horizontal or vertical directions and then shuffle these parts (detailed algorithm can be found in Algorithm \ref{alg_ps}). Both the two images will be fed into our detector, producing dense score maps at different scale levels. (In FCOS, there are 5 levels, \ie, $v\in[1,\cdots,5]$.)

To improve the generalization performance of SSOD, we adopt the following regularization loss:
\begin{align}
L_{scale}&=\sum_{v=1}^{4}\|p^{v}[U_{d}]-p^{v+1}[U_{sp}]\|^{2}_{2},
\end{align}
\noindent
where $p^{v}[U_{*}]$ indicates the score map $p^{v}$ derived from image $U_{*}$. Since the downsampling ratio $r=2$, $p^{v}[U_{d}]$ has the same resolution as $p^{v+1}[U_{sp}]$, and they are constrained to be consistent. 

\textbf{Remarks:} The output dense score maps  reveal the uncertainty or the reliability of the predicted label for each pixel. The lower the score is, the higher the uncertainty that the pixel belongs to a foreground object. Data uncertainty has been widely used to indicate the data importance in previous works \cite{kendall2017uncertainties,kendall2018multi,wang2021data,heo2018uncertainty,chen2016weakly}. In this paper, we regularize the uncertainty consistency. Patch shuffle is used to reduce the dependency of foreground objects on their surrounding contexts, improving the model robustness to context variations. In addition, to ensure consistent outputs among scales, $L_{scale}$ is then defined to improve the model robustness to object scaling variations.

By far, all the components of our DSL have been described, and the overall pipeline is shown in Figure \ref{fig_pipeline}.

\begin{algorithm}[t]
\small
    \caption{Patch Shuffle}\label{alg_ps}
    Input: Unlabeled image $U$\;
    Output: Patch shuffled image $U_{p}$\;
    Initialization: $U^{0}=U$, total iteration number $J$\;
    \For{$j=0,\cdots,J-1$}{
        (1) Mode $m$: randomly select a mode from [`horizontal',`vertical']\;
        (2) Normalized size $s$: randomly generate $s$ from interval $[0,1]$\;
        (3) Crop $U^{j}$ into two parts based on mode $m$ and normalized size $s$\;
        (4) Shuffle the order of the two parts, and concatenate them into a new image $\hat{U}^{j}$\;
        (5) $U^{j+1}=\hat{U}^{j}$;
    }
    \vspace{-0.5em}
\end{algorithm}

\vspace{-0.5em}
\section{Experiments}
\textbf{Datasets \& Evaluation Metrics:} We conduct experiments on the popular object detection benchmarks, including MS-COCO \cite{lin2014microsoft} and PASCAL-VOC \cite{everingham2010pascal}. MS-COCO contains more than 118$k$ labeled images, and there are about 850$k$ instances from 80 classes. In addition, there are 123$k$ unlabeled images provided for semi-supervised learning. VOC07 contains 5,011 training images from 20 classes, while VOC12 has 11,540 training images.

On MS-COCO, we follow the settings in STAC \cite{sohn2020simple} and evaluate with both the protocols of Partially Labeled Data and Fully Labeled Data. The former randomly samples 1\%, 2\%, 5\% and 10\% of the training data as labeled data, and treats the remainder as unlabeled data. (For this protocol, we create 3 data folds and report the mean results over them.) The latter uses all the training data as labeled data and the additional unlabeled data as unlabeled samples. We adopt the mean average precision $AP_{50:90}$ (denoted by mAP) as the evaluation metric.

For experiments on PASCAL-VOC07, following STAC \cite{sohn2020simple}, we use the VOC07 training set as the labeled data, and the VOC12 training set or together with the images from the same 20 classes in MS-COCO (denoted by COCO20) as the unlabeled data. We adopt VOC default $AP_{50}$ metric and COCO default mAP metric as the evaluation metrics.

\begin{table*}[!t]
  \centering
  \caption{The mAP performance (\%) of competing methods on the MS-COCO \cite{lin2014microsoft} dataset. The used protocol is Partially Labeled Data. $\dag$ means that the method uses a larger batch size $32$ or $40$, and $\ddag$ indicates that strong augmentation is applied on the labeled data. Note that $\dag, \ddag$ are not the default settings in STAC\cite{sohn2020simple} but they will improve the performance of both supervised baseline and SSOD. `Supervised' means that only the corresponding labeled data are used for training, and this is set as the baseline for SSOD.}
  \vspace{-0.5em}
  \resizebox{0.8\linewidth}{!}{
    \begin{tabular}{c|c|c|c|c|c|c}
    \hline
    \multicolumn{2}{c|}{Methods} & Deployment   & 1\% & 2\% & 5\% & 10\% \\
    \hline\hline
    \multirow{8}[0]{*}{Anchor-based} & Supervised \cite{sohn2020simple} & Hard & 9.05 \small{± 0.16} & 12.70 \small{± 0.15} & 18.47 \small{± 0.22} & 23.86 \small{± 0.81} \\
          & CSD \cite{jeong2019consistency}   & Hard & 11.12 \small{± 0.15} & 14.15 \small{± 0.13} & 18.79 \small{± 0.13} & 24.50 \small{± 0.15} \\
          & STAC \cite{sohn2020simple}  & Hard & 13.97 \small{± 0.35} & 18.25 \small{± 0.25} & 24.38 \small{± 0.12} & 28.64 \small{± 0.21} \\
          & IT \cite{zhou2021instant}    & Hard & 16.00 \small{± 0.20} & 20.70 \small{± 0.30} & 25.50 \small{± 0.05} & 29.45 \small{± 0.15} \\
          & ISMT \cite{yang2021interactive} & Hard & 18.88 \small{± 0.74} & 22.43 \small{± 0.56} & 26.37 \small{± 0.24} & 30.53 \small{± 0.52} \\
          & Humble \cite{tang2021humble} & Hard & 16.96 \small{± 0.38} & 21.72 \small{± 0.24} & 27.70 \small{± 0.15} & 31.60 \small{± 0.28} \\
          & UB$^{\dag}$ \cite{liu2021unbiased}   & Hard & 20.75 \small{± 0.12} & 24.30 \small{± 0.97} & 28.27 \small{± 0.11} & 31.50 \small{± 0.10} \\
          & E2E$^{\dag\ddag}$ \cite{xu2021end}  & Hard & 20.46 \small{± 0.39} &   -    & 30.74 \small{± 0.08} & 34.04 \small{± 0.14} \\
          \hline\hline
    \multirow{2}[0]{*}{Anchor-free} & Supervised(Ours) &   Easy    & 9.53 \small{± 0.23} & 11.71 \small{± 0.26} & 18.74 \small{± 0.18} & 23.70 \small{± 0.22} \\
          & DSL(Ours)   &   Easy    & \textbf{22.03} \small{± 0.28} & \textbf{25.19} \small{± 0.37} & \textbf{30.87} \small{± 0.24} & \textbf{36.22} \small{± 0.18} \\
          \hline
    \end{tabular}%
    }
  \label{tab:sota1}%
  \vspace{-.5em}
\end{table*}%

\textbf{Implementation Details:} We adopt the popular anchor-free detector FCOS \cite{tian2019fcos} with ResNet50 \cite{he2016deep} as backbone, and FPN \cite{lin2014microsoft} as neck and dense heads. Images in MS-COCO are resized to have shorter edge 800, or 640 if the longer edge is less than 1,333. Images in PASCAL-VOC are resized to have shorter edge 600, or 480 if the longer edge is less than 1,000. For fair comparison, following \cite{sohn2020simple,liu2021unbiased}, in all experiments, random flip is used as weak augmentation, while strong augmentation includes random flip, color jittering and cutout. The iteration $J$ is set to 2 in Patch Shuffle. For training configurations, learning rate starts from 0.01 and is divided by 10 at 16 and 22 epochs. The max epoch is 24. $\alpha$ is set to $3$ and $1$ for the partially and fully labeled protocols, \emph{resp}, and $2.5$ for VOC. $\epsilon$ is set to $0.99$. For parameter $\tau_{2}^{k}$, we set it within the range $[0.25,0.35]$. All of our experiments are based on Pytorch \cite{pytorch} and MMDetection \cite{mmdetection}. We use 8 NVIDIA-V100 GPUs with 32G memory per GPU. For each GPU, we randomly sample 2 images from labeled set and unlabeled set with ratio 1:1.

\subsection{Comparison with State-of-the-Arts}
We compare the proposed DSL with existing SOTA methods that are based on anchor-based detectors such as Faster-RCNN\cite{ren2015faster} and SSD\cite{liu2016ssd}. The results are shown in Tables \ref{tab:sota1}, \ref{tab:sota2} and \ref{tab:sota3}.

From Table \ref{tab:sota1}, one can see that under the supervised setting of the Partially Labeled Data protocol in COCO, our anchor-free detector achieves similar baseline performance to those anchor-based detectors, \ie, $9.53$ vs. $9.05$, $11.71$ vs. $12.70$, $18.74$ vs. $18.47$ and $23.7$ vs. $23.86$ with 1\%, 2\%, 5\% and 10\% labeled data, respectively. This means that anchor-free and anchor-based SSOD models are comparable when partially labeled data are used. After applying the proposed DSL algorithm, the SSOD performance can be significantly and consistently improved over the baselines under all protocols. DSL outperforms all the competing methods by a large margin, demonstrating the effectiveness and superiority of our method.

We also conduct experiments following the Fully Labeled Data protocol of COCO. The results are shown in Table \ref{tab:sota2}. Since the reported performance of those supervised methods varies a lot in the original works, we report their results together with their baselines, and compare their relative performance improvements. From Table \ref{tab:sota2}, one can see that our DSL achieves the largest performance improvement, \ie, $3.6$ mAP gain. The results on PASCAL-VOC are listed in Table \ref{tab:sota3}. We can see that the proposed DSL also achieves significant performance improvements over the supervised baselines as well as all the compared methods.

\begin{table}[!t]
  \centering
  \caption{The mAP performance (\%) of competing methods on the MS-COCO \cite{lin2014microsoft} dataset. The used protocol is Fully Labeled Data.}\vspace{-0.5em}
  \resizebox{0.8\linewidth}{!}{
    \begin{tabular}{c|c|c|c}
    \hline
    \multicolumn{2}{c|}{Methods} & Deployment & 100\% \\
    \hline\hline
    \multicolumn{1}{c|}{\multirow{4}[0]{*}{Anchor-based}} & STAC\cite{sohn2020simple}  & Hard  & 37.6$\stackrel{1.6}{\longrightarrow}$39.2 \\
          & ISMT\cite{yang2021interactive}  & Hard  & 37.8$\stackrel{1.8}{\longrightarrow}$39.6 \\
          & UB$^{\dag}$\cite{liu2021unbiased}    & Hard  & 40.2$\stackrel{1.1}{\longrightarrow}$41.3 \\
          & E2E$^{\dag\ddag}$\cite{xu2021end}   & Hard  & 40.9$\stackrel{3.6}{\longrightarrow}$44.5 \\
          \hline\hline
    Anchor-free & DSL(Ours)   & Easy  & \textbf{40.2$\stackrel{\textbf{3.6}}{\longrightarrow}$43.8} \\
    \hline
    \end{tabular}%
    }
  \label{tab:sota2}%
  \vspace{-.5em}
\end{table}%

\begin{table*}[!tbp]
  \centering
  \caption{The results (\%) of competing methods on the PASCAL-VOC \cite{everingham2010pascal} dataset. The performances are evaluated on the VOC07 test set.}
  \vspace{-.5em}
  \resizebox{0.8\linewidth}{!}{
    \begin{tabular}{c|c|c|cc|cc}
    \hline
    \multicolumn{2}{c|}{\multirow{2}[0]{*}{Methods}} & \multirow{2}[0]{*}{Deployment} & \multicolumn{2}{c|}{Unlabeled: VOC12} & \multicolumn{2}{c}{Unlabeled: VOC12 + COCO20} \\
    \multicolumn{2}{c|}{} &       & $AP_{50}$  & $AP_{50:90}$ & $AP_{50}$  & $AP_{50:90}$ \\
    \hline\hline
    \multicolumn{1}{c|}{\multirow{6}[0]{*}{Anchor-based}} & Supervised\cite{sohn2020simple} & Hard  & 72.75 & 42.04 & 72.75 & 42.04 \\
          & CSD\cite{jeong2019consistency}   & Hard  & 74.7  & -      & 75.1  & - \\
          & STAC\cite{sohn2020simple}  & Hard  & 77.45 & 44.64 & 79.08 & 46.01 \\
          & IT\cite{zhou2021instant}    & Hard  & 78.3  & 48.7  & 79    & 49.7 \\
          & ISMT\cite{yang2021interactive}  & Hard  & 77.23 & 46.23 & 77.75 & 49.59 \\
          & UB$^{\dag}$\cite{liu2021unbiased}    & Hard  & 77.37 & 48.69 & 78.82 & 50.34 \\
          \hline\hline
    \multirow{2}[0]{*}{Anchor-free} & Supervised(Ours) & Easy  & 69.6  & 45.9  & 69.6  & 45.9 \\
          & DSL (Ours)   & Easy  & \textbf{80.7}  & \textbf{56.8}  & \textbf{82.1}  & \textbf{59.8} \\
          \hline
    \end{tabular}%
    }\vspace{-.5em}
  \label{tab:sota3}%
\end{table*}%

In summary, the results in Tables \ref{tab:sota1}, \ref{tab:sota2} and \ref{tab:sota3} all demonstrate the effectiveness of our DSL method. It is worth mentioning that the proposed DSL is much easier to be deployed in real applications due to its negligible pre/post-processing costs compared to anchor-based methods, showing the great potential values of the anchor-free SSOD algorithm.

\subsection{Ablation Studies}
To better understand how the proposed DSL works, we conduct a series of ablation studies under the MS-COCO 10\% labeled data protocol.

\begin{table}[!tbp]
  \centering
  \caption{Effectiveness of each component of the proposed DSL method. `+' means training by the proposed method.} 
  \vspace{-0.5em}
    \begin{tabular}{lc}
    \hline
    Methods & mAP \\
    \hline\hline
    Supervised & 23.7 \\
    \hline
    + AF & 32.2 \\
    + MetaNet & 32.5 \\
    + AT & 34.5 \\
    + Patch-Shuffle & 34.9 \\
    + $L_{scale}$ & 36.2 \\
    \hline
    \end{tabular}%
  \label{tab:ablation}%
    \vspace{-.5em}
\end{table}%

\textbf{Effectiveness of each component.} The contributions of different components of DSL are listed in Table \ref{tab:ablation}. From this table, one can see that by using AF, the performance can be significantly improved from $23.7$ to $32.2$ mAP, which has already surpassed most SOTA methods shown in Table \ref{tab:sota1}. By adopting the MetaNet to refine the foreground pseudo-labels, the performance can be further improved to $32.5$. By applying AT to encourage the stability and quality of the pseudo-labels, the performance is further improved to $34.5$ mAP. Finally, by learning from shuffled patches and constraining the consistency among image scales, the overall model becomes more robust and exhibits higher accuracy, \ie, $36.2$ mAP. The ablation studies in Table \ref{tab:ablation} verify the effectiveness of each module in DSL.

\textbf{Ablation studies on AF.} Table \ref{tab:ai} shows the ablation studies on our AF strategy. In order to demonstrate the importance of multiple thresholds, we experiment with a single threshold strategy as reference, where instances are regarded as foreground if their scores are above the threshold and background otherwise. One can see that the single threshold strategy cannot achieve satisfactory performance. The best result is only $30.7$ mAP when the threshold is set to $0.2$, indicating that there are many instances being wrongly defined by a single threshold. In contrast, by using our multi-level thresholds strategy, \ie, AF, the performance can be significantly improved: even by using a fixed $\tau_{2}^{k}$=$0.3$, the result can be improved to $36.0$ mAP; and when the adaptive $\tau_{2}^{k}$ is used for each class, it can be further improved to $36.2$ mAP, showing the effectiveness and importance of our AF strategy.

\textbf{Ablation studies on AT.} From Table \ref{tab:at}, one can see that layer aggregation (LA) achieves higher performance gain than EMA because it considers the fine-grained relationships across layers, while EMA just simply aggregates layer-wise parameters independently so that the relationships between layers can be harmed. In addition, by employing both EMA and LA, our AT can further improve the performance to $36.2$ mAP. This implies that aggregations over parameters and layers are actually complementary.

\textbf{Ablation studies on loss weight $\alpha$.} From Table \ref{tab:alpha}, one can see that the performance peaks around $\alpha=3$. A too large weight such as $\alpha=4$ will give the model too many chances to employ the unlabeled images in training, and hence reduce the stability of the model.

\begin{table}[!t]
  \centering
  \caption{Ablation studies on Adaptive Filtering.}
    \vspace{-0.5em}
  \resizebox{1\linewidth}{!}{
    \begin{tabular}{c|cccc|ccc|c}
    \hline
    \multirow{2}[0]{*}{Methods} & \multicolumn{4}{c|}{Single threshold} & \multicolumn{3}{c|}{AF(fixed $\tau_{2}^{k}$)} & \multirow{2}[0]{*}{AF} \\
        &0.05  & 0.1   & 0.2   & 0.3   & 0.2   & 0.3   & 0.4   &  \\
          \hline
    mAP & 27.1 & 28.8  & 30.7  & 27.5  & 34.3  & 36.0  & 35.6    & \textbf{36.2} \\
    \hline
    \end{tabular}%
  \label{tab:ai}%
  }
  \vspace{-.5em}
\end{table}%

\begin{table}[!t]
  \centering
  \caption{Ablation studies on Aggregated Teacher. `LA' means layer aggregation.}
    \vspace{-0.5em}
    \begin{tabular}{c|c|cc|c}
    \hline
    Methods & No teacher & \emph{+} EMA   & \emph{+} LA   & AT \\
    \hline
    mAP   & 33.0    & 34.1  & 35.0    & \textbf{36.2} \\
    \hline
    \end{tabular}%
  \label{tab:at}%
  \vspace{-.5em}
\end{table}%

\begin{table}[!t]
  \centering
  \caption{Ablation studies on loss weight $\alpha$ for unlabeled data. `fail' means that the training loss will easily get to `nan'.}
    \vspace{-0.5em}
    \begin{tabular}{c|cccc}
    \hline
    $\alpha$ & 1     & 2     & 3     & 4 \\
    \hline\hline
    mAP   &    33.9   &   35.4    & \textbf{36.2}  & fail \\
        \hline
    \end{tabular}%
  \label{tab:alpha}%
  \vspace{-.5em}
\end{table}%

\textbf{Discussions.} In anchor-based SSOD, the negative/ignorable instances have been implicitly handled by label assigner and sampler, and we only need to consider how to recall the foreground instances via a threshold. In contrast, in anchor-free SSOD the multi-level pseudo-labels should be explicitly considered due to the pixel-wise gradient propagation. This can be demonstrated by our AF strategy as in Table \ref{tab:ai}. Moreover, without the help of predefined anchors for scale variances, FPN \cite{lin2014microsoft} with a dense head has been widely used in anchor-free detectors to address the scaling issue. Thus $L_{scale}$ can be generally adopted and regarded as a default trick in anchor-free SSOD, and this is verified to be effective in Table \ref{tab:ablation}. In summary, most of our techniques are proposed by considering the special characteristics of anchor-free detectors, and our work in this paper makes the first step towards anchor-free SSOD.

\section{Conclusion}
In this paper, we made the first attempt, to the best of our knowledge, to bridge the gap between SSOD and anchor-free detector, and developed a DSL based SSOD method. The DSL was built upon several novel techniques, such as Adaptive Filtering, Aggregated Teacher and uncertainty regularization. Our experiments showed that the proposed DSL outperformed the state-of-the-art SSOD methods by a large margin on both COCO and VOC datasets. It is expected our work can inspire more and in-depth explorations on anchor-free SSOD methods.

{\small
\bibliographystyle{ieee_fullname}
\bibliography{egbib}
}

\end{document}